\RequirePackage{fix-cm}
\documentclass[twocolumn, natbib]{svjour3}          
\smartqed  
\usepackage{graphicx}

\usepackage{times}
\usepackage{epsfig}
\usepackage{amsmath}
\usepackage{amssymb}
\usepackage{subfig}
\usepackage{tabularx,booktabs,multirow,makecell}
\usepackage{caption}
\usepackage[dvipsnames]{xcolor}
\usepackage{enumitem}

\usepackage[pagebackref=true,breaklinks=true,citecolor=blue,linkcolor=blue,letterpaper=true,colorlinks,bookmarks=false]{hyperref}
\usepackage{xcolor}

\newcommand{\specialcell}[2][c]{\begin{tabular}[#1]{@{}c@{}}#2\end{tabular}}

\newcommand{\revise}[1]{#1}
\newcommand{\comment}[1]{}

\begin{document}\sloppy

\title{Learning Contrastive Representation for Semantic Correspondence
}


\author{Taihong Xiao$^1$ \and Sifei Liu$^2$ \and Shalini De Mello$^2$ \and Zhiding Yu$^2$ \and Jan Kautz$^2$ \and Ming-Hsuan Yang$^{1,3}$  
}

\authorrunning{Taihong Xiao~\etal} 

\institute{
	Taihong Xiao \at
	\email{txiao3@ucmerced.edu}           
	\and
	Sifei Liu \at
	\email{sifeil@nvidia.com}
    \and
    Shalini De Mello \at
	\email{shalinig@nvidia.com}
	\and
	Zhiding Yu \at
	\email{zhidingy@nvidia.com}
	\and
	Jan Kautz \at
	\email{jkautz@nvidia.com}
	\and
	Ming-Hsuan Yang \at
	\email{mhyang@ucmerced.edu}
	\\
$^1$University of California, Merced, CA, USA\\
$^2$Nvidia, Santa Clara, CA, USA\\
$^3$Yonsei University, Seoul, Korea
}

\date{Received: date / Accepted: date}

\maketitle

\begin{abstract}
Dense correspondence across semantically related images has been extensively studied, but still faces two challenges: 1) large variations in appearance, scale and pose exist even for objects from the same category, and 2) labeling  pixel-level dense correspondences is labor intensive and infeasible to scale. 
%
%
Most existing methods focus on designing various matching modules using fully-supervised ImageNet pretrained networks.
%
On the other hand, while a variety of self-supervised approaches are proposed to explicitly measure image-level similarities, correspondence matching the pixel level remains under-explored. 
In this work, we propose a multi-level contrastive learning approach for semantic matching, which does not rely on any ImageNet pretrained model. 
We show that image-level contrastive learning is a key component to encourage the convolutional features to find correspondence between similar objects, while the performance can be further enhanced by regularizing cross-instance cycle-consistency at intermediate feature levels.
Experimental results on the PF-PASCAL, PF-WILLOW, and SPair-71k benchmark datasets demonstrate that our method performs favorably against the state-of-the-art approaches.
The source code and trained models will be made available to the public. 

\end{abstract}


\section{Introduction}
  
\begin{figure*}[htb] 
	\centering 
    \includegraphics[width=0.6\textwidth]{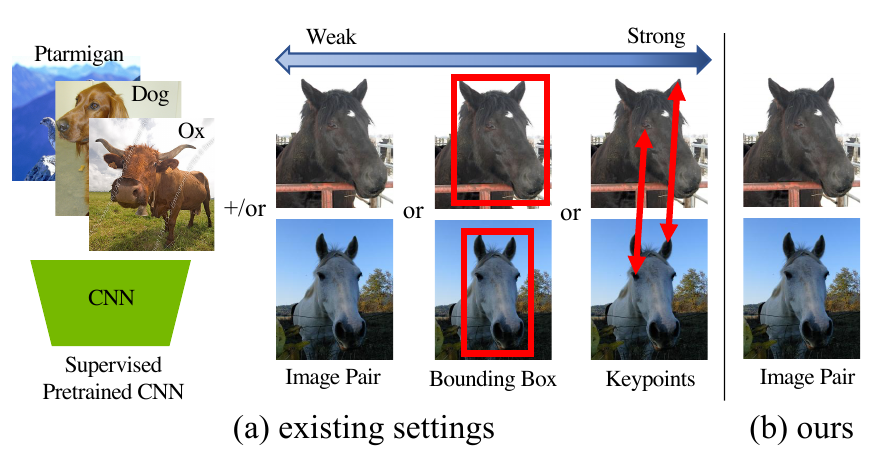}
    \caption{Visual comparison of existing and our settings in terms of the used supervision. Most existing works rely on supervised pretrained CNN and/or pairwise supervision. From weak to strong, the pairwise supervision includes image pair, bounding box, and keypoints. In contrast, our method only requires that image pairs belong to the same category for training. } 
    \label{fig:supervision} 
\end{figure*}

\begin{table*}[htb]
	\caption{\revise{Supervisory signals used by evaluated methods. The last column indicates that the validation image pairs with keypoint annotations are used for model selection. 
	%
	%
	Note that knowing keypoint correspondence between two images indicates that they form an image pair of the same category, and the bounding boxes of these objects can thus be extracted.}}
    \label{table:supervision-comparison} 
    \centering
	\begin{tabular}{l|c|ccc|c}
        \Xhline{1pt}
        \multicolumn{1}{c|}{\multirow{2}[0]{*}{Methods}} & \multicolumn{1}{c|}{Supervised} & \multicolumn{3}{c|}{Training} & \multicolumn{1}{c}{Validation} \\
              &  \multicolumn{1}{c|}{Pretrained CNN}   & \multicolumn{1}{c}{Image Pair}  & \multicolumn{1}{c}{Bounding Box} & \multicolumn{1}{c|}{Keypoints} &  Keypoints \\
        \hline
        DCTM~\citep{kim2017dctm} & \checkmark &       &   &  \checkmark   & \checkmark \\
        SC-Net~\citep{Han2017SCNetLS} & \checkmark &       &   &  \checkmark   & \checkmark \\
        Weakalign~\citep{Rocco2018EndtoEndWS}  & \checkmark & \checkmark    &       &       & \checkmark \\
        RTNs~\citep{Kim2018RecurrentTN}  & \checkmark & \checkmark    &       &       & \checkmark\\
        NC-Net~\citep{Rocco2018NeighbourhoodCN}  & \checkmark & \checkmark     &       &       & \checkmark \\
        DCC-Net~\citep{Huang2019DynamicCC} & \checkmark & \checkmark  &       &       & \checkmark \\
        HPF~\citep{Min2019HyperpixelFS}   & \checkmark &       &       &       & \checkmark \\
        DHPF~\citep{Min2020LearningTC}  & \checkmark & \checkmark  &       &       & \checkmark \\
        SF-Net~\citep{Lee2019SFNetLO} & \checkmark &       & \checkmark &       & \checkmark \\
        PARN~\citep{Jeon2018PARNPA} & \checkmark &       & \checkmark &       & \checkmark \\
        FCSS~\citep{Kim2019FCSSFC} & \checkmark &       & \checkmark &       & \checkmark \\
        SCOT~\citep{Liu2020SemanticCA}  & \checkmark &       &       &       & \checkmark \\
        ours  &       & \checkmark    &       &       &  \\
        \Xhline{1pt}
    \end{tabular}
\end{table*}

Semantic correspondence is one of the fundamental problems in computer vision with many applications in object recognition~\citep{Duchenne2011AGK,Liu2011SIFTFD}, image editing~\citep{Dale2009ImageRU}, semantic segmentation~\citep{Kim2013DeformableSP}, and scene parsing~\citep{Zhou2015FlowWebJI}, to name a few. 
The goal is to establish dense correspondences across images containing the objects or scenes of the same category. 
%
For example, as shown in Fig.~\ref{fig:supervision}(a), such dense correspondence is established across two horse images, where the semantically similar keypoints such as eyes or ears of different horses are matched.
%
However, this task is extremely challenging as different objects usually appear with distinctive appearances caused by variations in shapes, lighting, poses and scales.
Classic approaches~\citep{Bristow2015DenseSC,Hur2015GeneralizedDS,Kim2013DeformableSP,Liu2011SIFTFD,Taniai2016JointRO,Yang2014DAISYFF} 
determine correspondence matching via hand-crafted features such as SIFT~\citep{LoweDavid2004DistinctiveIF}, DAISY~\citep{Tola2010DAISYAE} and HOG~\citep{Dalal2005HistogramsOO}.
More recently, deep CNN based approaches \citep{Choy2016UniversalCN,Han2017SCNetLS,Jeon2018PARNPA,Kanazawa2016WarpNetWS,Kim2019FCSSFC,Novotn2017AnchorNetAW,Rocco2017ConvolutionalNN,Rocco2018EndtoEndWS,Seo2018AttentiveSA,Zhou2016LearningDC} have achieved significant improvements, by exploiting hierarchical image features that provide rich semantic features that are invariant to intra-class variations.

It is challenging to develop fully-supervised approaches for learning pixel-level matching as a large number of image pairs with detailed pixel correspondence annotations are required.   
To alleviate this issue, several methods exploit weakly-supervised information, e.g., object bounding boxes or foreground masks~\citep{Jeon2018PARNPA,Kim2019FCSSFC,Zhou2015FlowWebJI,Zhou2016LearningDC,Lee2019SFNetLO}, for this task. 
Nevertheless, it still entails significant amount of labor to annotate these labels for a large scale dataset. 
As a trade-off between model performance and labeling effort, several methods~\citep{Rocco2018EndtoEndWS,Kim2018RecurrentTN,Rocco2018NeighbourhoodCN,Huang2019DynamicCC,Min2020LearningTC} leverage image class labels as weak supervisory signals. 
For example, we can obtain image pairs from the same object or scene category, which provides weak supervision for learning the correspondence. 
However, existing methods predominantly rely on effective universal feature representations in the first place, which are often obtained by supervised ImageNet pretraining. 
Benefiting from large-scale labeled images, a supervised pretrained network can extract image-level discriminative features for semantic correspondence. 

Recent advances~\citep{Hjelm2019LearningDR,Chen2020ASF,He2020MomentumCF,Grill2020BootstrapYO} in self-supervised representation learning exploit contrastive loss functions to construct image-level discriminative features from unlabeled image data.  
%
%
Their goal is to push object representations of various views of the same object closer while pulling those of different objects further apart by learning to perform an image instance discrimination task.
%
The learned representations by these contrastive learning methods have achieved significant performance gains for numerous pretext tasks such as video object segmentation~\citep{Pathak2017LearningFB} and tracking~\citep{wang2015unsupervised}.  
However, learning a fine-grained representation at part- or pixel-level has not been well studied, especially for semantic correspondence without using ImageNet pretrained models. 
Some methods~\citep{Wang2020DenseCL,Pinheiro2020UnsupervisedLO} have generalized image-level contrastive learning to dense contrastive learning, but do not establish cross-instance correspondence. 
As shown in Fig.~\ref{fig:supervision} and summarized in Table~\ref{table:supervision-comparison}, existing approaches have used different levels of supervisory signals, while our model is learned based on weak supervision, i.e., image pairs of the same category,  without resorting to fully-supervised pretrained ImageNet networks.

To learn a generalizable representation for semantic correspondence, we develop a multi-level contrastive representation learning method in this paper. 
%
We show that by applying the contrastive learning framework~\citep{He2020MomentumCF} merely on the image level, the mid-level convolutional features can capture local correspondences between similar objects reasonably well. 
%
The results suggest that to learn good representations for high-level tasks (e.g., object recognition), lower-level features are enforced to learn correct correlations at a fine-grained level as well.
We embed a pixel-level contrastive learning scheme into the same network to further obtain fine-grained feature representation learning by enforcing  cross-instance cycle consistency regularization at intermediate feature levels. 
Given a pair of images with semantically similar objects, we track selected pixels from the source to the target images and then back to the source via an affinity matrix. 
\revise{We then enforce that those selected pixels map back to their original locations via the cross-instance cycle consistency constraint.} 
Essentially, cycle consistency is equivalent to pixel-level contrastive learning, where the path of each pixel can be considered as either positive or negative depending on whether it forms a cycle~\citep{Jabri2020SpaceTimeCA}. 
To avoid trivial solutions, we use a self-attention module to localize the foreground object and apply a group of augmentations based on the computed attention map. 
The main contributions of this work can be summarized as follows:
\begin{itemize}
	\item We pose a new weakly-supervised semantic matching problem by relaxing the strong dependency on supervised ImageNet pretrained models and removing the validation ground truth used for model selection. 
	\item We develop a multi-level contrastive learning framework where image-level contrastive learning generates object-level discriminative representations and pixel-level contrastive learning further facilitates fine-grained representations at region- or pixel-level to improve dense semantic correspondence performance.
	\item We propose a cross-instance cycle consistency regularization to learn a discriminative local feature at the pixel-level without dense ground truth by leveraging different objects of the same category in different images instead of the same object in self-augmented images or video sequences.
	\item We demonstrate that the proposed model performs favorably against the state-of-the-art method on three datasets with comprehensive ablation studies on components of our framework.
\end{itemize}

\section{Related Work}

\subsection{Semantic Correspondence}


%
Numerous methods exploit hierarchical features from deep models pretrained on the ImageNet dataset to infer semantic correspondence.  
In these approaches, semantic matching is formulated as a geometric alignment task and addressed via the self-supervised learning framework where training image pairs and ground truth are synthesized based on in-plane transformations~\citep{Jeon2018PARNPA,Kanazawa2016WarpNetWS,Kim2018RecurrentTN,Rocco2017ConvolutionalNN,Rocco2018EndtoEndWS,Han2017SCNetLS,Seo2018AttentiveSA}. 
On the other hand, weak-supervision signals, such as image-level labels~\citep{Rocco2018EndtoEndWS,Kim2018RecurrentTN,Rocco2018NeighbourhoodCN,Huang2019DynamicCC,Min2020LearningTC}, bounding boxes~\citep{Jeon2018PARNPA,Kim2019FCSSFC,Zhou2015FlowWebJI,Zhou2016LearningDC,Lee2019SFNetLO}, and keypoints~\citep{Han2017SCNetLS} have also been used for semantic correspondence. 
In contrast, the proposed method does not require fully supervised ImageNet pretrained networks and only utilizes image-level supervision, i.e., image pairs of same category, to determine semantic correspondence from images.

In addition, numerous approaches freeze the ImageNet pretrained backbone model and determine the network structure of other modules or hyper-parameters by exploiting ground truth keypoint correspondences of a small validation set as supervisory signals~\citep{Min2019HyperpixelFS,Liu2020SemanticCA,Huang2019DynamicCC,chen2018deep}.
%
%
\cite{Min2019HyperpixelFS} leverage a small number of relevant features selected from early to late layers of a convolutional neural network as well as beam search to construct hyperpixel layers, and use regularized Hough matching (RHM) to infer semantic correspondence efficiently. 
%
\cite{Liu2020SemanticCA} pose semantic matching as an optimal transport problem and solve it using the Sinkhorn's algorithm~\citep{Sinkhorn1967DiagonalET}. 
%
These two methods require only supervision from keypoints of a small validation set during the beam search stage to select feature layers from a pretrained deep model (e.g., ResNet-50 and ResNet-101). 
%
%
We note that keypoint ground truth from a validation set provides strong supervision for model or hyper-parameter selection, which is not weaker than either pixel-level training supervision or supervised ImageNet pretrained model weights. 
In contrast, our method does not require any keypoint correspondence supervision from a validation dataset as the proposed cross-instance cycle consistency regularization can be used for self-supervised learning.

\subsection{Self-Supervised Representation Learning}

Self-supervised representation learning methods can be classified into three categories: generative, inductive (predictive), and contrastive. 
%
Generative models, e.g.,~\citep{Vincent2008ExtractingAC,Oord2016ConditionalIG,Oord2016PixelRN}, maximize the likelihood of observed data based on various probabilistic formulations and representation, e.g., auto-encoders and convolutional neural networks. 
%
In contrast, inductive methods predict some known information from the data as there exist ground truths for comparison, which is typically carried out by applying various augmentation techniques, including image rotations prediction~\citep{Gidaris2018UnsupervisedRL}, spatial configuration of cropped patches~\citep{Doersch2015UnsupervisedVR}, video colorization~\citep{Zhang2016ColorfulIC}, video sequence order sorting~\citep{Misra2016ShuffleAL}, and jigsaw puzzle solving~\citep{Noroozi2016UnsupervisedLO}. 

Recently, numerous contrastive learning methods, such as Deep InfoMax~\citep{Hjelm2019LearningDR}, SimCLR~\citep{Chen2020ASF}, MoCo~\citep{He2020MomentumCF} and BYOL~\citep{Grill2020BootstrapYO}, have shown that effective representation models can be constructed without supervision, with performance comparable to fully supervised ones. 
%
%
%
%
However, image-level contrastive learning may not produce optimal features for semantic correspondence, where the association is built between pixels. 
A few methods~\citep{Wang2020DenseCL,Pinheiro2020UnsupervisedLO,Jabri2020SpaceTimeCA} propose to learn dense visual representations where the association is required between pixels, which are direct extensions of image-level instance contrastive learning methods (e.g., MoCo~\citep{He2020MomentumCF}). 
They compute contrastive losses between different views of the {\it same} instance, e.g., either via self-augmented images~\citep{Wang2020DenseCL,Pinheiro2020UnsupervisedLO}, or from videos of the same instance~\citep{Jabri2020SpaceTimeCA}. 
However, their learned representations only captures the instance-level information, and thus cannot be used to establish the pixel-level cross-instance semantic correspondence.
%
%
In contrast, we propose a multi-level cross-instance pixel contrastive learning method by leveraging {\it different} instances via a cyclic framework to infer semantic correspondence.

\revise{\cite{kang2020pixel} leverage the pixel-level cycle association of source and target pixel pairs across two different domains for contrastive representation learning. 
However, it requires knowing the
pixel-level categories in advance to construct positive/negative samples for contrastive learning. 
In contrast, the proposed method does not entail any pixel-level supervisory signals to form positive/negative pixel pairs. 
On the other hand,~\cite{xie2021propagate} propose a joint pixel-level and instance-level contrastive learning framework. 
It differs from our approach in constructing positive and negative samples. 
The pixel-level contrastive learning in \citep{xie2021propagate} is carried out through pixels within the same instance, i.e., the corresponding pixels in two views of the same image are considered as positive pairs. 
Nevertheless, the proposed model establishes pixel-level contrastive learning across different instances. 
Extensive experimental results show that the proposed cross-instance pixel-level contrastively learned representation can better handle challenges caused by large appearance and pose distinctions across different instances.}

\subsection{Temporal Correspondence}
Our work is related to temporal correspondence learning, for which a number of self-supervised approaches~\citep{Vondrick2018TrackingEB, Oord2018RepresentationLW,Li2019JointtaskSL,Wang2019LearningCF,Jabri2020SpaceTimeCA} have been explored. 
\cite{Li2019JointtaskSL} propose to learn temporal correspondence by joining region-level localization and pixel-level matching through a shared inter-frame affinity matrix.
%
In \citep{Jabri2020SpaceTimeCA}, the correspondence learning problem is cast as link prediction in a space-time graph constructed from a video and long-range cycle consistency is exploited to learn temporal correspondence. 
Unlike these methods, we target a more challenging task for establishing correspondence across semantically similar instances with significant appearance and pose distinctions.

\subsection{Cycle Consistency}

Cycle consistency has been widely used as a constraint in numerous vision tasks. 
For example, in the context of image-to-image translation~\citep{zhu2017unpaired}, or face attributes editing~\citep{DBLP:conf/bmvc/ZhouXYFHH17,xiao2018dna,Xiao_2018_ECCV}, exploiting cycle consistency enables learning the mapping between different image domains from unpaired data. 
In the context of optical flow estimation, computing the forward and backward consistency~\citep{meister2018unflow,liu2019ddflow} can be used to effectively infer occluding pixels for learning optical flow. 

For learning correspondence, the cycle consistency constraint is formulated in various ways. 
\cite{Zhou2016LearningDC} construct a cross-instance loop between real and synthetic images and establish cross-instance correspondence through 3D CAD rendering.
SC-Net~\citep{Han2017SCNetLS} establishes a differentiable flow field by computing feature similarities while considering background clutter and proposes flow consistency loss between forward and backward flow fields.
\cite{zhou2015multi}~propose to optimize joint matching of multiple images via rank minimization by translating cycle consistency into positive semi-definiteness and low-rankness constraints. 
On the other hand, \cite{chen2018deep} exploit forward-backward consistency and transitivity consistency constraints to enforce geometrically plausible predictions. 
In contrast, our cycle consistency regularization is established via self augmentations to learn finer-grained pixel-level representations. 






\section{Proposed Method}

\begin{figure*}
    \centering
    \includegraphics[width=0.99\textwidth]{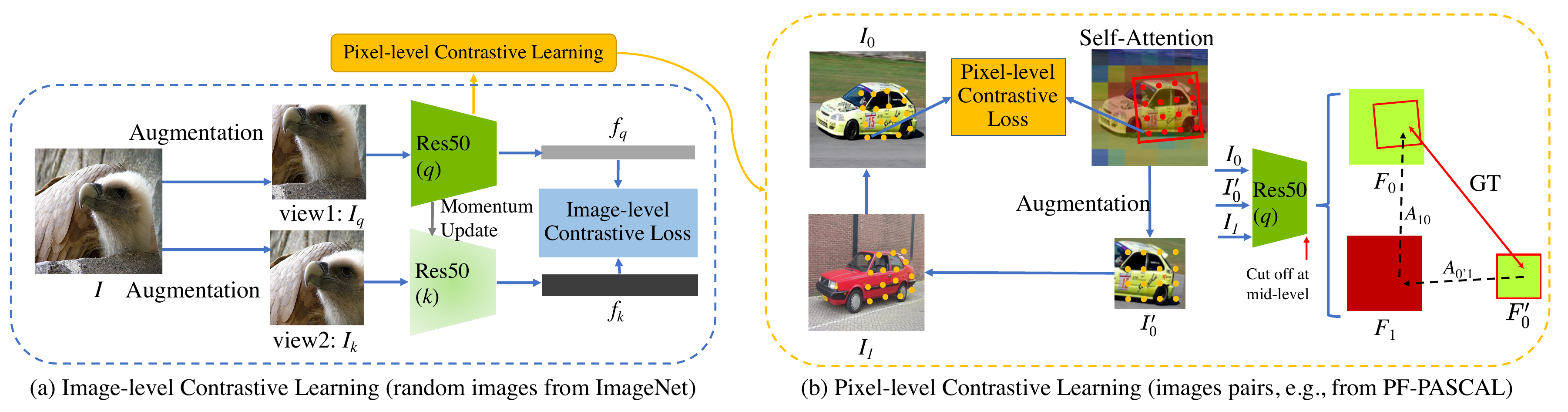}
    \caption{Overview of our framework. The dark green trapezoid denotes the backbone network $q$ shared by (a) image-level and (b) pixel-level contrastive learning. In (a), we use random images from ImageNet as the inputs, and the feature from its last layer for image-level contrastive learning, where the image-level contrastive loss is defined as~\eqref{eq:image-constrastive-loss}. In (b), we use image pairs of the same category (e.g., from PF-PASCAL) as inputs and extract their mid-level features for pixel-level contrastive learning, where the pixel-level contrastive loss is defined as~\eqref{eq:pixel-contrastive-loss}.}
    \label{fig:framework}
\end{figure*}

In this section, we introduce the proposed contrastive representation learning model for semantic correspondence. 
As shown in Fig.~\ref{fig:framework}, the overall framework consists of image-level and pixel-level contrastive learning modules. 
%
We briefly describe the image-level contrastive learning schemes in Section~\ref{sec:image-level-contrastive-learning}, introduce the proposed pixel-level contrastive learning method in Section~\ref{sec:pixel-level-contrastive-learning} and our implementation details in Section~\ref{sec:implementation-pipeline}.
%

\subsection{Image-level Contrastive Learning}
\label{sec:image-level-contrastive-learning}


Image-level contrastive learning \citep{He2020MomentumCF,Chen2020ASF,Grill2020BootstrapYO} aims to extract object-level discriminative features in a self-supervised manner. 
It learns representations from object-centric images by minimizing the distance between two different views of an object generated through different augmentation methods. 
In addition, numerous methods~\citep{He2020MomentumCF,Chen2020ASF} use negative samples from different images and maximize the distance between  positive and negative observations. 
In this work, we exploit negative examples in a way similar to the MoCo method~\citep{He2020MomentumCF}.
%
Image-level contrastive learning can be considered as a dictionary look-up task. 
We use a dynamic queue of size $K$ to store a set of encoded keys $\{f_{1}, f_{2}, \ldots\}$, among which a single positive key $f_k$ matches with $f_q$. 
For a query $f_q$, its positive key $f_k$ encodes a different view of the same image, while the negative keys encode the views of different images. 
Thus, the image-level contrastive loss is defined as follow:
\begin{equation}\label{eq:image-constrastive-loss}
    L_{q} = -\log\frac{\exp(f_q \cdot f_k / \tau) }{\exp(f_q \cdot f_k / \tau) + \sum_{i\neq k}\exp(f_q \cdot f_{i} / \tau)},
\end{equation}
where $\tau$ denotes a temperature hyper-parameter. 
Both the query network $q$ and the key network $k$ share the same architecture, but $q$ is updated based on the image-level contrastive loss using back-propagation while $k$ is updated with a momentum function:
\begin{equation}\label{eq:momentum_update_k}
	\theta_k \leftarrow m\theta_k + (1-m) \theta_q,
\end{equation}
where $\theta_q$ and $\theta_k$ are parameters of $q$ and $k$, and $m\in[0,1)$ denotes the momentum coefficient.

For image-level contrastive learning, each image is regarded as a unique category.
%
Therefore the function for image-level contrastive learning approaches~\eqref{eq:image-constrastive-loss} is essentially equivalent to a $(K+1)$-way classification task, where $K$ is the size of the queue. 
It facilitates learning image-level discriminative representations without using any image labels. 
%
%
We show in our experiments (see Table~\ref{tab:loss-datasets}) that even without additional objective functions, the learned image-level representation models can perform semantic correspondence matching well (see Fig.~\ref{fig:vis-comparison}).

\subsection{Pixel-level Contrastive Learning}
\label{sec:pixel-level-contrastive-learning}

To learn fine-grained representation models, we propose a pixel-level contrastive learning method explicitly for semantic correspondence.
As shown in Fig.~\ref{fig:framework}(b), a pair of images $I_0$ and $I_1$ containing different objects of the same category are provided as the inputs. 
We then obtain an attention map highlighting the foreground object region via the self-attention module. 
Based on the attention map, we obtain $I'_0$ by applying a series of spatial data augmentations to $I_0$, including random horizontal flipping, rotation, and cropping. 
We use the same encoder network $q$, shared with the image-level module, to extract features for $I_0$, $I_1$, and $I'_0$, which are denoted by $F_0$, $F_1$ and $F'_0$, respectively, as shown in Fig.~\ref{fig:framework}(b)). 
We extract feature maps from the middle layers of ResNet50 instead of the feature vector from the last layer to learn the pixel-level representation.
We introduce each sub-module in the following sections.  

\paragraph{Self-attention module.} 
We introduce a self-attention module to avoid including pixels from background regions. 
For the given source image $I_0$, we extract the feature $f_0\in\mathbb{R}^{C}$ after the last layer and the feature $K_0\in\mathbb{R}^{C\times H\times W}$ before the global average pooling layer, where $H$ and $W$ denote the size of the feature map. 
As the feature map $K_0$ keeps the spatial location information of pixels in $I_0$, we compute cosine similarity between  normalized $f_0$ and all feature vectors in the normalized attention map $K_0$. 
%
\revise{The self-attention module is used to exclude background pixels when applying a series of augmentation to $I_0$ to obtain $I'_0$. 
The random image regions are cropped near the pixels using the largest cosine similarity. 
As such, the background pixels are less likely to be contained in the augmented image $I'_0$, thus eliminating the number of pixel paths starting from the background for more effective cross-instance cycle consistency.}

\paragraph{Associating pixels via affinity.}
%
Given a pair of images $I_0$ and $I_1$ and their mid-level features $F_0\in\mathbb{R}^{C\times H_0 W_0}$ and $F_1\in\mathbb{R}^{C\times H_1 W_1}$, we use the correlation matrix $R_{01}\in\mathbb{R}^{H_0 W_0\times H_1 W_1}$ to represent the pixel-level similarity as: 
\begin{equation}\label{eq:correlation_matrix}
    R_{01} = F_0^\top F_1.
\end{equation}
%
To ensure a one-to-one mapping, the correlation matrix needs to be sparse.
However, it is challenging to model a sparse matrix in a deep neural network. 
Therefore, we relax this constraint and encourage the correlation matrix to be sparse by normalizing each row with the softmax function.
As such, the similarity score distribution can be peaky, and only a few pixels with high similarity in the source image are matched to each point in the target image. 
The affinity matrix is defined by:
\begin{equation}\label{eq:affinity_matrix}
    A_{01} = \mathrm{softmax}(R_{01} / t),
\end{equation}
where $t$ is the temperature hyperparameter controlling how peaky the normalized distribution is. 
The affinity matrix enjoys several good properties: 1) The summation over each row is unity since softmax is applied to the row dimension. 
2) The multiplication of two affinity matrices results in an affinity matrix. 
3) The affinity matrix can be used to trace the corresponding pixel locations of feature map $F_0$ in the target feature map $F_1$, defined by $P_{01} = A_{01}G_1$, where $G_1\in\mathbb{R}^{H_1 W_1\times 2}$ is a vectorized pixel location map (i.e., each element denotes its horizontal and vertical positions).

\paragraph{Pixel-level guidance.} We carry out the proposed pixel-level contrastive learning by formulating the correspondence as a graph, where the nodes are image pixels and the edges are weighted by the similarities between their features in the latent space. 
Starting from $I'_0$ to $I_1$ and then back to $I_0$, we track corresponding pixels by computing two affinity matrices $A_{0'1}$ and $A_{10}$ following~\eqref{eq:affinity_matrix} using their mid-layer features. 
We enforce cross-instance cycle consistency by requiring pixels from $I'_0$ to be mapped back to where they are located in $I_0$, through a different image $I_1$. 
The affinity matrix along the path of cycle can be simply obtained through multiplication, $\bar{A}_{0'0} = A_{0'1}A_{10}\in\mathbb{R}^{H'_0W'_0\times H_0W_0}$, where $H$ and $W$ denote the feature size. 
Each element in the cycle affinity matrix $\bar{A}_{0'0}$ depicts a pixel path from $I'_0$ to $I_0$ passing through $I_1$. 
We predict corresponding pixel locations of the patch image $I'_0$ in the source image $I_0$ by $P = \bar{A}_{0'0} G_0$.
As the ground truth correspondence $\hat{P}$ between $I'_0$ and $I_0$ is known, the pixel-level contrastive loss is defined as:
\begin{equation}\label{eq:pixel-contrastive-loss}
    L_p = ||P - \hat{P}||_2.
\end{equation}
We note that \eqref{eq:pixel-contrastive-loss} is a variant of the contrastive objective w.r.t. pixels from a pair of images. 
Unlike the image-level loss \eqref{eq:image-constrastive-loss} that explicitly pushes and pulls on the positive and negative pairs, the cycle loss matches a group of ``starting'' and ``ending'' pixels.
Specifically, all pixels sampled in the walking path are considered positive, while the other pairs are negative. 
The feature representation is learned to be pixel-wisely discriminative when the affinity matrix is enforced to be ``peaky'' in each row.


\paragraph{Information entropy regularization.} Information entropy loss~\citep{Min2020LearningTC} can also be used as a regularization term to learn affinity by encouraging more distinctive correspondences. 
We empirically find that training the network by the pixel-level contrastive loss with the information entropy loss can further improve the performance of semantic correspondence. 
With the correlation matrix $R\in\mathbb{R}^{H_0W_0\times H_1W_1}$ as defined in \eqref{eq:correlation_matrix}, we compute the correlation entropy as:
\begin{equation}\label{eq:correlation_entropy}
    H(R) = -\frac{1}{H_0W_0} \sum_{i=1}^{H_0W_0}\sum_{j=1}^{H_1W_1}\phi(R)_{ij}\log\phi(R)_{ij},
\end{equation}
where $\phi(\cdot)$ denotes row-wise $\ell_1$ normalization, and $\phi(R)_{ij}$ denotes the $(i,j)$-th elements of $\phi(R)$. 
As lower correlation entropy indicates more distinctive correspondence between two images, we encourage low entropy for the image pair $I_0$ and $I_1$ by using the following information entropy loss:
\begin{equation}\label{eq:information-entropy-loss}
    L_r = H(R_{01}) + H(R_{10}),
\end{equation}
where $R_{01}$ and $R_{10}$ are correlation matrices between the source and target images.
\revise{We use correlation matrices $R$ instead of affinity matrices $A$ to compute the information entropy loss. 
%
The reason is that each row in $A=\mathrm{softmax}(R/t)$ has been normalized with summation of each row to 1, and thus all entries would have been normalized twice if we use $\phi(A)$.}

\subsection{Implementation Details}
\label{sec:implementation-pipeline}

\paragraph{Pretraining and training.} We use ResNet50~\citep{He2016DeepRL} as the backbone network for feature extraction. 
For image-level contrastive learning, we use the feature from the last layer of ResNet50, while the feature from an intermediate layer (e.g., the feature after $13^\mathrm{th}$ Res-block) for pixel-level contrastive learning.
Following the same hyperparameter settings of MoCo~\citep{He2020MomentumCF}, we pretrain the backbone network using image-level contrastive learning on the ImageNet dataset for initialization. 
Then we train the backbone network with a combination of image-level and pixel-level contrastive losses along with the information entropy regularization,
\begin{equation}
    L = \lambda_p L_p + \lambda_q L_q + \lambda_r L_r,
\end{equation}
where $\lambda_p$, $\lambda_q$, and $\lambda_r$ are weight coefficients for the pixel-level contrastive loss~\eqref{eq:pixel-contrastive-loss}, image-level contrastive loss~\eqref{eq:image-constrastive-loss} and the information entropy regularization~\eqref{eq:information-entropy-loss}, respectively. 
Note that for image-level contrastive learning, we do not need a 1000-class labeled dataset, but just random images. Neither do we need that for pixel-level contrastive learning. In contrast, only image pairs from the same category are required for pixel-level contrastive learning.
%
\revise{We empirically set $\lambda_p=0.0005$, $\lambda_q=1$, $\lambda_r=0.001$, and the temperature $t=0.0007$ in~\eqref{eq:affinity_matrix}.}

\paragraph{Validation: beam search without ground truth correspondence.}\label{sec:beam-search}
To utilize the learned representation model for semantic correspondence, we need to choose multi-layer features for testing, which is also called {\it hyperpixel} construction~\citep{Min2019HyperpixelFS}.
Existing methods~\citep{Min2019HyperpixelFS,Liu2020SemanticCA} use the beam search algorithm to find the optimal subset of deep convolutional layers according to performance on the validation split.
However, the existing beam search method requires accessing the ground truth correspondence of the validation set. 
%
%
To relax the dependency on validation annotations, we perform beam search over all convolutional layers of a given deep model by using the proposed pixel-level contrastive loss as the performance indicator. 
A lower pixel-level contrastive loss means a better layer combination. 
%
\revise{The validation process is only used for hyperpixel selection and not for other hyper-parameter selection.}

\begin{figure*}[htp]
\centering
\def\picwidth{0.24\textwidth}
\def\tabwidth{0.22\linewidth}

\begin{tabular}{m{0.09\linewidth}m{0.22\linewidth}m{0.2\linewidth}m{0.24\linewidth}m{0.12\linewidth}}
 & Raw & OT & OT + RHM & GT \\
\end{tabular}
\rotatebox{90}{\;\;Baseline} 
\includegraphics[width=\picwidth]{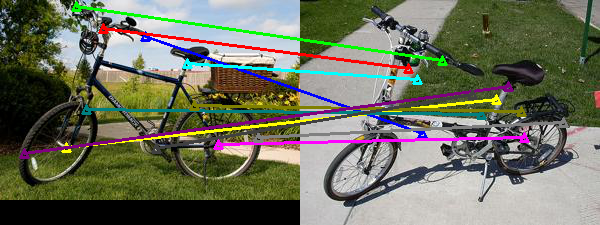} 
\includegraphics[width=\picwidth]{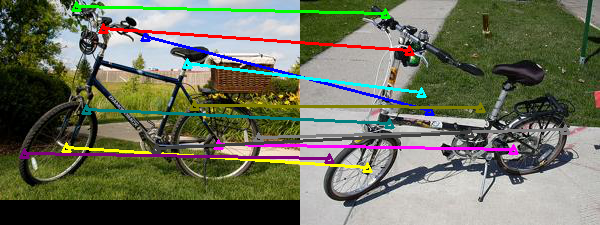} 
\includegraphics[width=\picwidth]{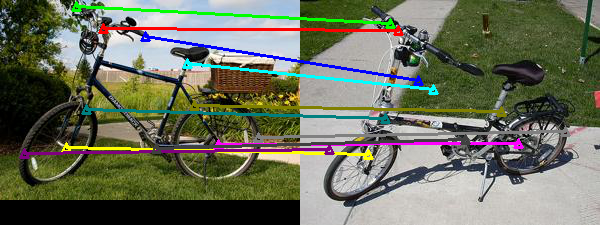} 
\includegraphics[width=\picwidth]{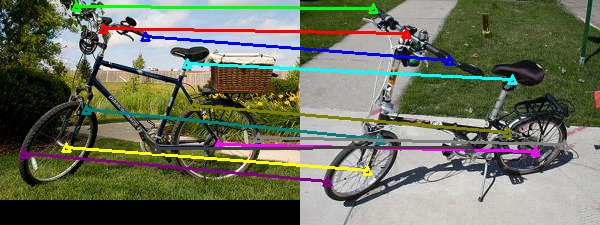} \\
\rotatebox{90}{\;\;\;Ours} 
\includegraphics[width=\picwidth]{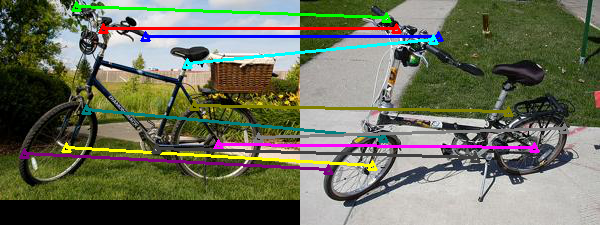} 
\includegraphics[width=\picwidth]{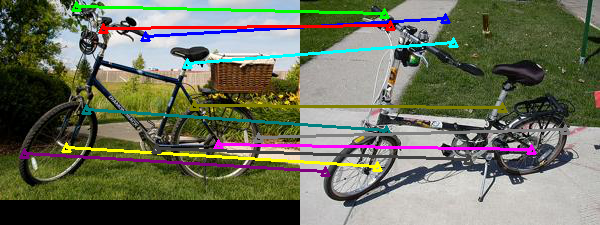} 
\includegraphics[width=\picwidth]{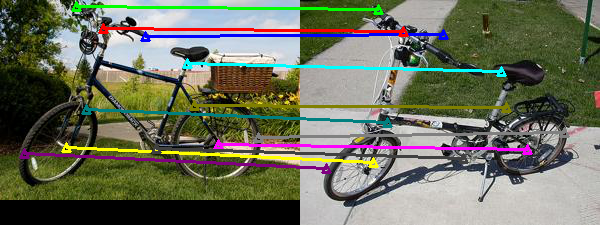} 
\includegraphics[width=\picwidth]{./extra/vis-ours-GT-022} \\
\vspace{2mm}
\rotatebox{90}{\;Baseline} 
\includegraphics[width=\picwidth]{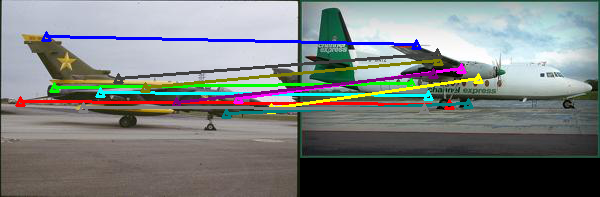} 
\includegraphics[width=\picwidth]{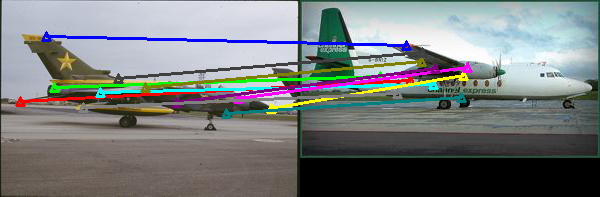} 
\includegraphics[width=\picwidth]{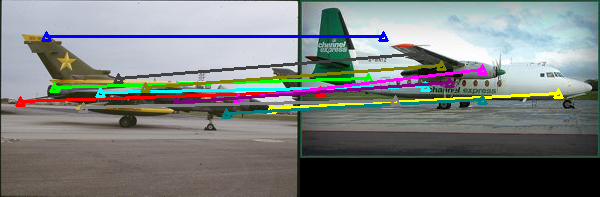} 
\includegraphics[width=\picwidth]{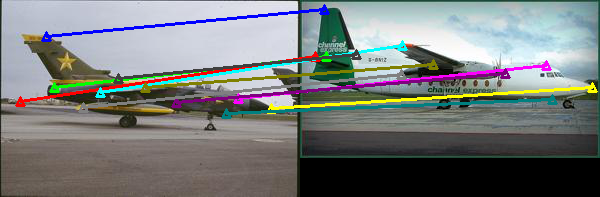} \\
\rotatebox{90}{\;\;\;Ours} 
\includegraphics[width=\picwidth]{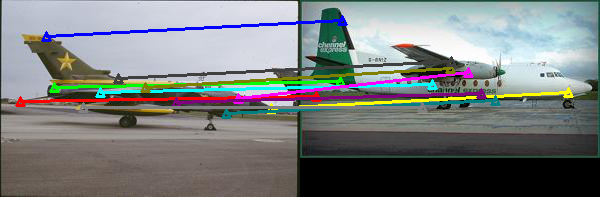} 
\includegraphics[width=\picwidth]{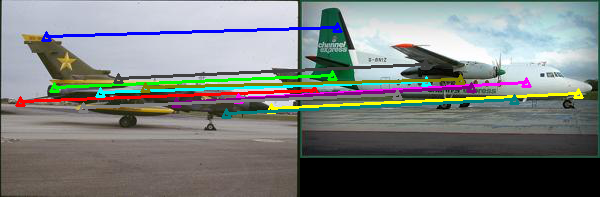} 
\includegraphics[width=\picwidth]{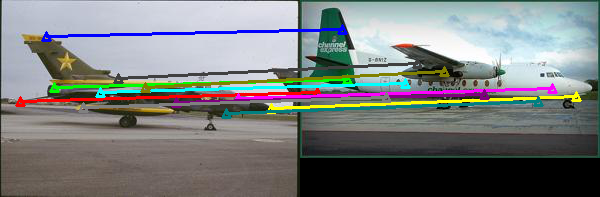} 
\includegraphics[width=\picwidth]{./extra/vis-ours-GT-012} \\
\vspace{2mm}
\rotatebox{90}{\;Baseline} 
\includegraphics[width=\picwidth]{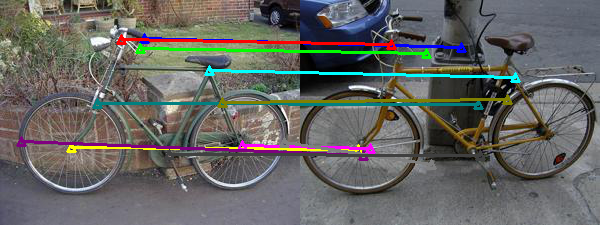} 
\includegraphics[width=\picwidth]{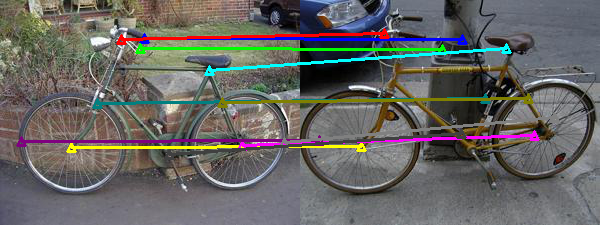} 
\includegraphics[width=\picwidth]{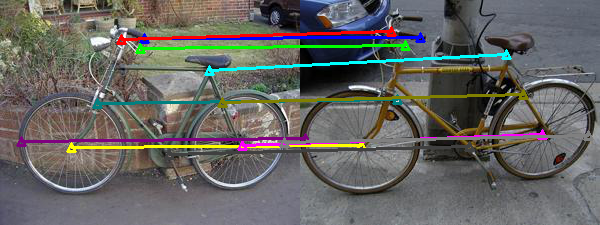} 
\includegraphics[width=\picwidth]{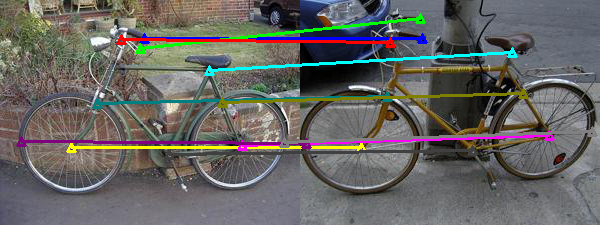} \\
\rotatebox{90}{\;\;\;Ours} 
\includegraphics[width=\picwidth]{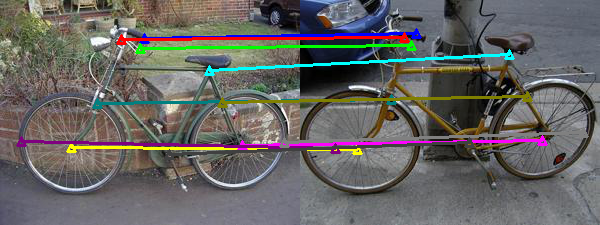} 
\includegraphics[width=\picwidth]{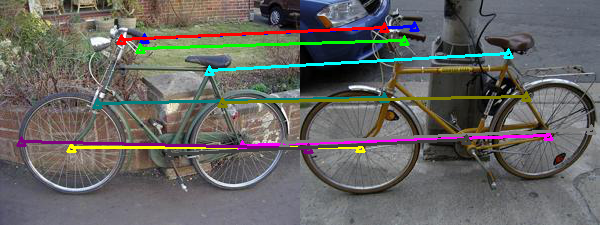} 
\includegraphics[width=\picwidth]{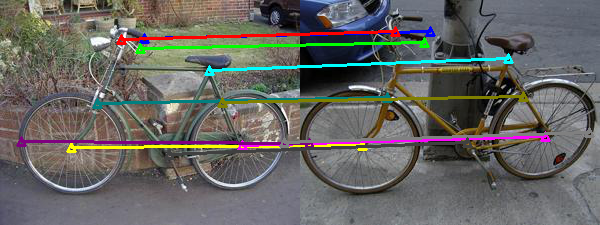} 
\includegraphics[width=\picwidth]{./extra/vis-ours-GT-032} \\
\vspace{2mm}
\rotatebox{90}{\;Baseline} 
\includegraphics[width=\picwidth]{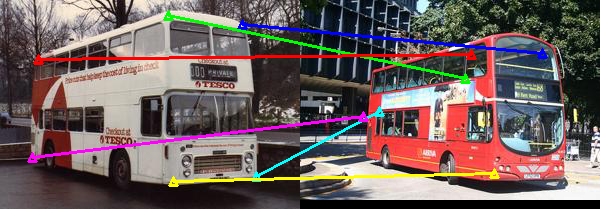} 
\includegraphics[width=\picwidth]{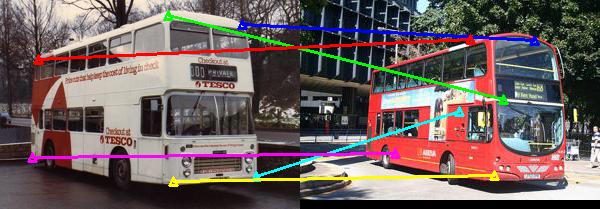} 
\includegraphics[width=\picwidth]{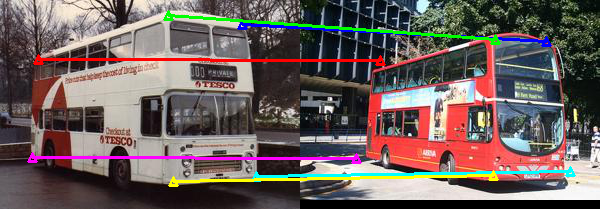} 
\includegraphics[width=\picwidth]{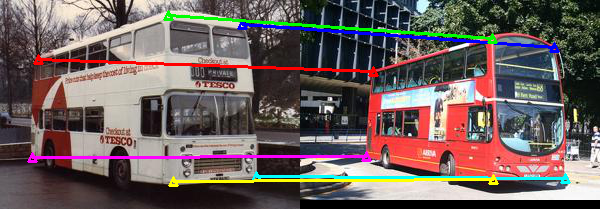} \\
\rotatebox{90}{\;\;\;Ours} 
\includegraphics[width=\picwidth]{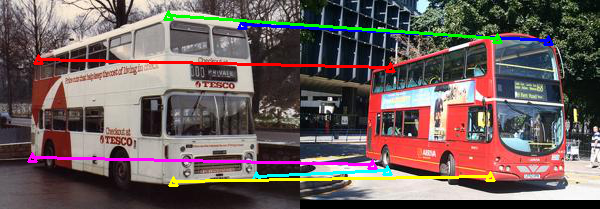} 
\includegraphics[width=\picwidth]{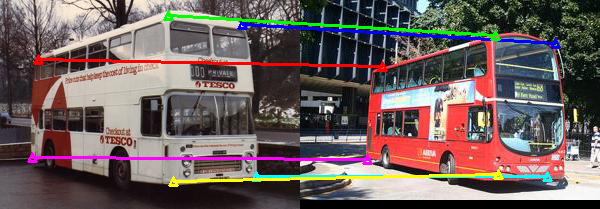} 
\includegraphics[width=\picwidth]{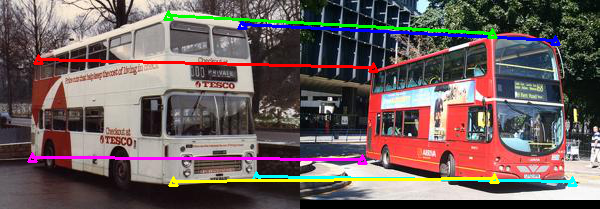} 
\includegraphics[width=\picwidth]{./extra/vis-ours-GT-092} \\
\vspace{2mm}
\rotatebox{90}{\;Baseline} 
\includegraphics[width=\picwidth]{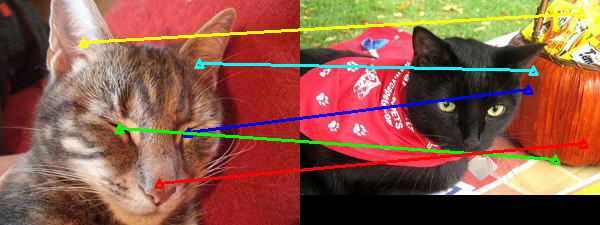} 
\includegraphics[width=\picwidth]{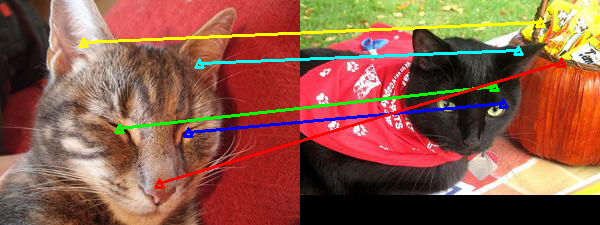} 
\includegraphics[width=\picwidth]{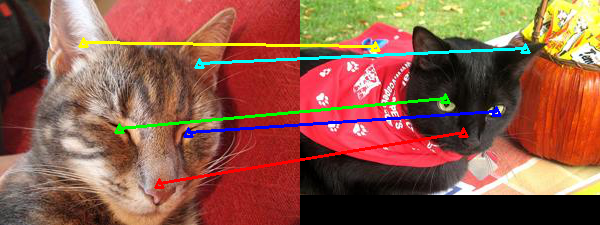} 
\includegraphics[width=\picwidth]{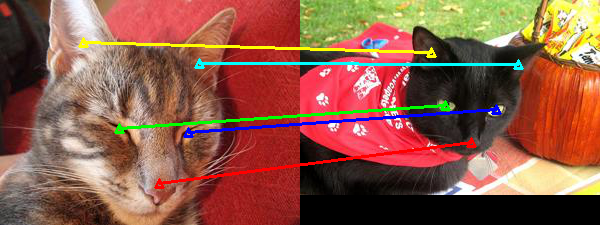} \\
\rotatebox{90}{\;\;\;Ours} 
\includegraphics[width=\picwidth]{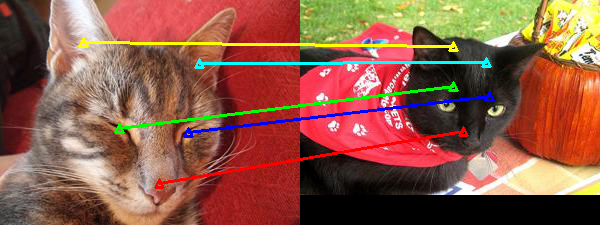} 
\includegraphics[width=\picwidth]{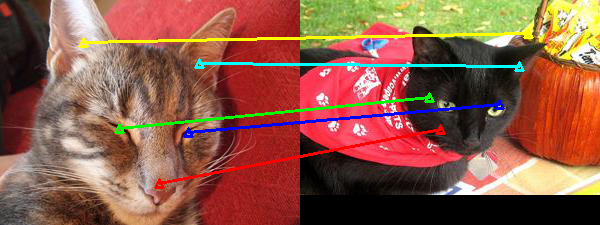} 
\includegraphics[width=\picwidth]{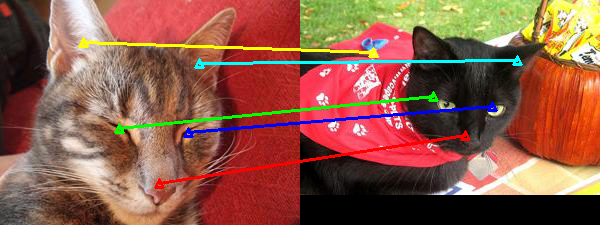} 
\includegraphics[width=\picwidth]{./extra/vis-ours-GT-132} \\
\vspace{2mm}
\rotatebox{90}{\;Baseline} 
\includegraphics[width=\picwidth]{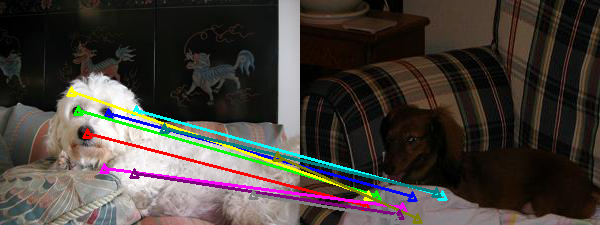} 
\includegraphics[width=\picwidth]{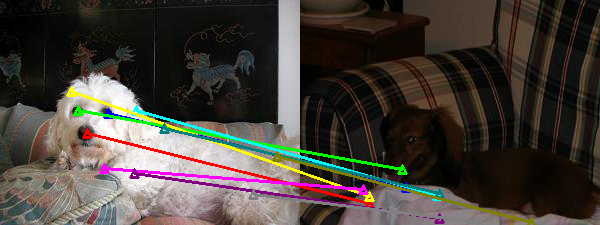} 
\includegraphics[width=\picwidth]{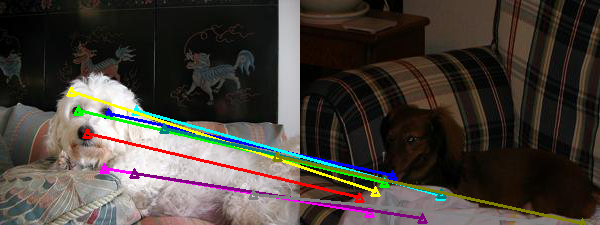} 
\includegraphics[width=\picwidth]{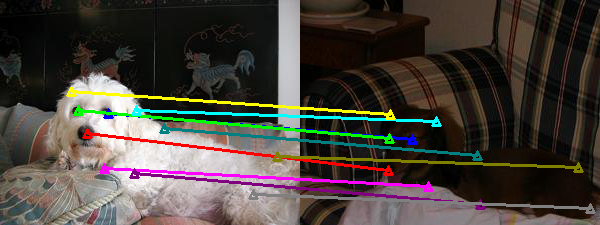} \\
\rotatebox{90}{\;\;\;Ours} 
\includegraphics[width=\picwidth]{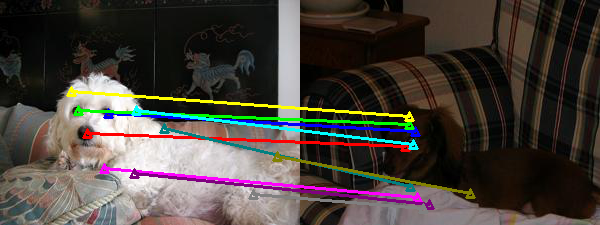} 
\includegraphics[width=\picwidth]{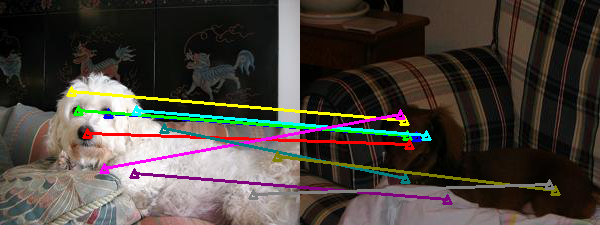} 
\includegraphics[width=\picwidth]{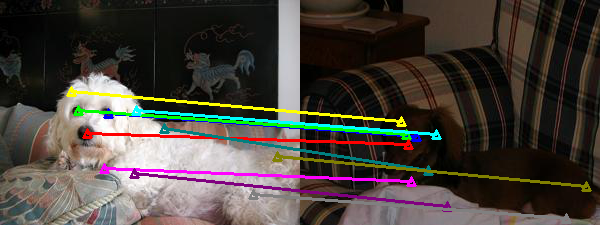} 
\includegraphics[width=\picwidth]{./extra/vis-ours-GT-182} \\
\caption{Visualization of the baseline and our method on the PF-PASCAL dataset. The first three columns show the correspondence predictions made based on the affinity matrix (Raw), transport matrix (after OT) and voting matrix (after OT + RHM). The last column is the ground truth correspondence. Different colors indicate different keypoint matches.}
\label{fig:vis-comparison}
\end{figure*}

\paragraph{Matching process.} Given a pair of images, we extract their features and compute the affinity matrix based on the selected hyperpixels. 
Similar to the steps in SCOT~\citep{Liu2020SemanticCA}, we first perform the optimal transport (OT) to relax the affinity matrix and obtain the transport matrix between two images. 
By viewing matching problem as an optimal transport (OT) problem, we perform image matching at a global perspective compared to matching pixels independently.
The OT problem can be solved using Sinkhorn's algorithm~\citep{Sinkhorn1967DiagonalET}. Consequently, the many-to-one matching issue can be alleviated.
Then we employ the regularized Hough matching (RHM)~\citep{Min2019HyperpixelFS} as the post-processing step to obtain the voting matrix and determine final keypoint correspondences. 
The RHM method further improves the matching accuracy by enforcing geometric consistency by reweighting the matching score in the Hough space. 
We carefully analyse the effectiveness of these post-processing steps and the results can be found in Table~\ref{tab:beamsearch-OT-RHM} and Fig.~\ref{fig:vis-comparison}.


\begin{table*}[htbp]
  \centering
  \caption{Evaluation results on the PF-PASCAL and PF-WILLOW datasets. We compare our method against others with three different thresholds. Based on whether we add the entropy loss Eq.~\eqref{eq:information-entropy-loss} or not, there are two variants of method, denoted by w/o and w/ entropy in the table. Numbers in bold indicate the best performance. 
  }
    \begin{tabular}{l|c|c|c|c|c|c}
	\Xhline{1pt}
    \multicolumn{1}{c|}{\multirow{2}[0]{*}{Methods}} & \multicolumn{3}{c|}{PF-PASCAL ($\alpha_{img}$)} & \multicolumn{3}{c}{PF-WILLOW ($\alpha_{bbox}$)} \\
	\cline{2-7}
          & \multicolumn{1}{l|}{PCK@0.05} & \multicolumn{1}{l|}{PCK@0.10} & \multicolumn{1}{l|}{PCK@0.15} & \multicolumn{1}{l|}{PCK@0.05} & \multicolumn{1}{l|}{PCK@0.10} & \multicolumn{1}{l}{PCK@0.15} \\
 	 \Xhline{1pt}
     Weakalign~\citep{Rocco2018EndtoEndWS} & 30.83 & 60.90 & 76.49 & 27.98 & 56.29 & 72.42 \\
     NC-Net~\citep{Rocco2018NeighbourhoodCN} & 41.34 & 62.89 & 72.14 & -- & -- & -- \\
     DCC-Net~\citep{Huang2019DynamicCC} & 	45.28 & 72.26 & 82.43 & 37.29 & 65.01 & 78.10 \\
     DHPF~\citep{Min2020LearningTC} & 45.64 & 73.96 & 85.69 & 43.09 & 69.01 & 82.14 \\
     SCOT~\citep{Liu2020SemanticCA} & 44.30 & 71.20 & 83.80 & 37.20 & 62.40 & 75.90 \\
     \hline
     Ours(w/o entropy) & {\bf 51.00} & {\bf 77.10} & {\bf 88.50} & 40.10 & 66.40 & 80.30  \\
     Ours(w/ entropy) & 50.70 & 76.10 & 85.80 & {\bf 43.10} & {\bf 69.80} & {\bf 82.40} \\
     \Xhline{1pt}
    \end{tabular}%
  \label{tab:comparison-sota}%
\end{table*}%

\section{Experiments and Analysis}

We propose a new benchmark setting specifically for learning semantic correspondence without using any supervised ImageNet pretrained network or validation ground truth.
For all the evaluated methods, we use the ResNet50 model~\citep{He2016DeepRL} as the backbone network for feature extraction, in a way similar to the self-supervised pretraining scheme of~\citep{He2020MomentumCF}. 
We also use a unified standard for the validation step by performing model selection without resorting to using the ground truth keypoint correspondence of the validation image pairs.

In Section~\ref{sec:dataset-metric}, we first describe the datasets and evaluation metrics.
Then, we present results as per the new benchmark setting by comparing our method with state-of-the-art methods in Section~\ref{sec:comparison-sota}. 
In Section~\ref{sec:ablation-study}, we conduct comprehensive ablation studies to analyze different components and variations of our model.


\subsection{Datasets and Evaluation Metrics}
\label{sec:dataset-metric}

We evaluate the proposed method on three benchmark datasets: PF-PASCAL~\citep{Ham2018ProposalFS}, PF-WILLOW~\citep{ham2016proposal} and SPair-71k~\citep{min2019spair}. 
The PF-PASCAL dataset contains 1,351 image pairs from 20 object categories of the PASCAL VOC~\citep{Everingham2014ThePV} database, and the PF-WILLOW dataset contains 900 image pairs of 4 object categories. 
The SPair-71k dataset is a more challenging large-scale dataset consisting of keypoint-annotated 70,958 image pairs from 18 categories with diverse view-point and scale variations.
We carry out pixel-level contrastive learning on one of these three datasets and image-level contrastive learning method on the ImageNet dataset~\citep{deng2009imagenet}.

The PF-PASCAL dataset consists of 1300 image pairs with keypoint annotations of 20 object classes. 
Each pair of images in the PF-PASCAL dataset share the same set of non-occluded keypoints. 
We divide the dataset into 700 training pairs, 300 validation pairs, and 300 testing pairs following~\citep{Min2020LearningTC,Liu2020SemanticCA}. 
The image pairs for training/validation/testing are distributed proportionally to the number of image pairs of each object class. 

For quantitative evaluation, we adopt the widely-used percentage of correct keypoints (PCK) metric, which counts the number of correctly predicted keypoints given a fixed threshold. 
Given a predicted keypoint $k_{pr}$ and the ground truth keypoint $k_{gt}$, the prediction is considered correct if:
\begin{equation}\label{eq:pck}
    d(k_{pr}, k_{gt}) \le \alpha_\tau \cdot \max(w_\tau, h_\tau),
\end{equation}
where $d(\cdot, \cdot)$ is the Euclidean distance, $w_\tau$ and $h_\tau$ are the width and height of either an entire image or object bounding box, i.e., $\tau\in\{\mathrm{img}, \mathrm{bbox}\}$, and $\alpha$ is a fixed threshold. 
We compute the final PCK by averaging the results from all testing image pairs.

\subsection{Evaluation against the State-of-the-art Methods}
\label{sec:comparison-sota}

Due to the full supervision and richer features of deeper neural networks (e.g., ResNet101), models that are pretrained with the large-scale labeled ImageNet have stronger representation strength. 
For fair comparisons, we \textit{re-implement all the evaluated methods} by initializing their models with the same MoCo~\citep{He2020MomentumCF} pretrained ResNet50 backbone network. 
We emphasize that the backbone networks weights of  NC-Net~\citep{Rocco2018NeighbourhoodCN}, DCC-Net~\citep{Huang2019DynamicCC} and DHPF~\citep{Min2020LearningTC} are all frozen during training as stated in the experimental sections in these papers. 
The SCOT method~\citep{Liu2020SemanticCA} does not involve any training. 
Instead, it adopts a fixed backbone and optimizes on top of the extracted features. 
Since all the methods are built on top of a fixed pretrained network, it is fair to evaluate all methods by replacing their backbones with the same MoCo pretrained ResNet50 model. 
Replacing these models with purely randomly initialized weights will degrade the performance.
%
%
Note that we specifically evaluate the methods that utilized the same level of weak supervision as ours, i.e., a group of matchable image pairs.

As aforementioned, the ground truth annotations of the validation set are utilized by numerous approaches, e.g., Weakalign~\citep{Rocco2018EndtoEndWS}, NC-Net~\citep{Rocco2018NeighbourhoodCN}, DCC-Net~\citep{Huang2019DynamicCC}, DHPF~\citep{Min2020LearningTC}, to determine the optimal hyper-parameters (neighborhood numbers, kernel size, etc.). 
For these method, we adopt the hyper-parameters in their original implementation in our re-implementations, because other randomly selected hyper-parameters would lead to lower performance.
%
In addition, we utilize the re-implemented SCOT~\citep{Liu2020SemanticCA} method as our baseline model, which shares a similar matching procedure, i.e., OT and RHM, but with features extracted from a fixed backbone network pretrained with the image-level contrastive loss.
%
For our implementation of SCOT, we remove the use of the validation set for beam search and replace it with the same strategy as proposed in Section~\ref{sec:beam-search}.
%
%

Table~\ref{tab:comparison-sota} shows improvements introduced by our proposed multi-level contrastive learning method. 
We also visualize the results of the re-implemented SCOT baseline model and our method in Fig.~\ref{fig:vis-comparison}. 
%
%
Our method generates more accurate and consistent matches than the baseline model (see more discussion of the reasons accounting for the improvement in the ablation study below).
%
%
%
The re-implemented SCOT baseline model shows that the intermediate features of the backbone network pretrained using image-level contrastive learning can capture the semantic correspondences well.
%
Furthermore, by embedding pixel-level contrastive learning into the network, the proposed model can adapt to the variations between different images at the pixel level, thereby generating more fine-grained matching results.  


\subsection{Ablation Studies}
\label{sec:ablation-study}
In this section, we analyze numerous components of the proposed method, including the image-level and pixel-level contrastive losses, beam search, OT and RHM, information entropy regularization, self-attention module, and the training layers. 
%
As each component plays an essential role in the model's performance, we conduct comprehensive ablation studies and discuss the contribution of each module.

\paragraph{Variants of image- and pixel-level contrastive learning.}\label{para:image-and-pixel-level-contrastive-learning-variants}
We analyze the proposed pixel-level contrastive learning with comparisons to other possible variants on the PF-PASCAL dataset, as summarized below and in Table~\ref{tab:loss-datasets}. 
For simplicity, we use IC and PC as the abbreviations of image-level and pixel-level contrastive learning variants:
\begin{enumerate}
    \item IC: We re-implement SCOT as our baseline model, where the backbone is pretrained via image-level contrastive learning, as aforementioned in Section~\ref{sec:comparison-sota}.
    \item IC (finetune): To verify whether the performance gain above the baseline model comes from the image pairs in the PF-PASCAL dataset or not, we finetune the baseline model via the image-level contrastive loss on both Image-Net and PF-PASCAL datasets. 
    \item IC + PC (selfcycle): In addition to image-level contrastive learning, we conduct pixel-level contrastive learning on self-augmented image pairs from ImageNet.
    \item IC + PC (align): We apply image- and pixel-level contrastive learning on both ImageNet and PF-PASCAL datasets, where the pixel-level contrastive loss of ImageNet is calculated on the self-augmented image pairs.
    \item IC (fix) + PC: To verify the effectiveness of multi-level contrastive learning, we fix the backbone pretrained model via image-level contrastive learning while training a side convolutional net branch combining features from multiple layers as input via the pixel-level contrastive loss. 
    The integrated feature following these additional layers is directly used for inference without applying beam search to the trained model.
    \item IC + PC (ours): We jointly conduct image-level contrastive learning (on ImageNet) and pixel-level contrastive learning (on PF-PASCAL).
\end{enumerate}

\begin{table}[tp]
  \centering
  \caption{Ablation study of the image- and pixel-level contrastive losses. All the numbers are evaluated at PCK@0.05 on the PF-PASCAL dataset. IC and PC are abbreviations of image-level contrastive learning and pixel-level contrastive learning. IN and PF stand for the ImageNet and PF-PASCAL datasets. Please refer to details of the ablation studies described in Section~\ref{para:image-and-pixel-level-contrastive-learning-variants} for details.}
    \begin{tabular}{l|c|c|c|c|c}
	\Xhline{1pt}
    \multicolumn{1}{c|}{\multirow{2}[0]{*}{Models}} & \multicolumn{2}{c|}{IC} & \multicolumn{2}{c|}{PC} & \multicolumn{1}{c}{\multirow{2}[0]{*}{\specialcell{PCK\\@0.05}}} \\
	\cline{2-5}
	  & IN & PF & IN & PF & \\
 	\Xhline{1pt}
 	IC & \checkmark &   &   &  & 44.3 \\
 	IC (finetune)  & \checkmark & \checkmark  &   &  & 43.7 \\
 	IC + PC (selfcycle)  & \checkmark &   & \checkmark &  & 38.1 \\
 	IC + PC (align)  & \checkmark &  \checkmark & \checkmark & \checkmark & 34.6 \\
 	IC (fix) + PC  &  &   &   & \checkmark & 39.5 \\
 	IC + PC (Ours) & \checkmark &   &   & \checkmark & {\bf 51.0} \\
 	\Xhline{1pt}
    \end{tabular}%
  \label{tab:loss-datasets}%
\end{table}%

As shown in Table~\ref{tab:loss-datasets}, the proposed IC + PC model outperforms the IC baseline method by a large margin (51.0\% v.s. 44.3\%). 
Either the effectiveness of pixel-level contrastive learning or using image pairs from the same category accounts for the performance gain. 
To better understand these effects, we conduct the IC (finetune) experiment. 
%
As it performs almost as well as the IC variant (43.7\% v.s. 44.3\%), the performance gain of the IC + PC model does not result from the weak supervision provided by the PF-PASCAL dataset.  
%
In other words, the proposed pixel-level contrastive learning method helps achieve the performance gain. 

Results of IC + PC (selfcycle) and IC + PC (align) show that using augmented image pairs from Image-Net for pixel-level contrastive learning negatively affects performance (38.1\% and 34.6\% v.s. 44.3\%). The underlying reason is that different views of the same object may lead to trivial solutions as such augmented image pairs cannot provide views of different objects with a large variation.
The result of IC (fix) + PC model is worse than that of the IC variant (39.5\% v.s. 44.3\%), which shows that model performance degrades without leveraging the image-level contrastive loss for updating the model weights. It also reflects that the image-level contrastive learning scheme is the cornerstone of the pixel-level contrastive learning scheme.
The representations learned via the image-level contrastive learning scheme lead to coarse correspondences, and they can be refined via the pixel-level contrastive learning scheme, leading to more accurate correspondences. 
%

\begin{table}[tp]
  \centering
  \caption{Ablation study of beam search, optimal transport (OT), and regularized Hough matching (RHM). The w/o GT in the second column denotes the proposed beam search by using the pixel-level contrastive loss as the indicator. The numbers in the third column indicate the res-block IDs used for hyperpixel construction. The columns of OT and RHM denote whether they were used in the testing phase. The numbers in the last column are evaluated at PCK@0.05 on the PF-PASCAL dataset.}
  \scalebox{1}{
    \begin{tabular}{l|c|c|c|c|c}
        \Xhline{1pt}
          & \specialcell{Beam\\ Search} & Hyperpixel & OT  & RHM & \specialcell{PCK\\@0.05} \\
        \Xhline{1pt}
        \multicolumn{1}{c|}{\multirow{6}[0]{*}{Baseline}} & w/o GT & (1,2,4,13,14) & \checkmark & \checkmark & 44.3 \\
          & w/o GT & (1,2,4,13,14)   & \checkmark &            & 20.9 \\
          & w/o GT & (1,2,4,13,14)   &            &            & 18.8 \\
          \cline{2-6}
          & w/ GT  & (3,11,12,13,15) & \checkmark & \checkmark & 52.4 \\
          & w/ GT  & (3,11,12,13,15) & \checkmark &            & 39.1 \\
          & w/ GT  & (3,11,12,13,15) &            &            & 34.3 \\
        \hline
        \multicolumn{1}{c|}{\multirow{6}[0]{*}{Ours}} & w/o GT & (2,12,13,15) & \checkmark & \checkmark & 51.0 \\
          & w/o GT & (2,12,13,15) & \checkmark &            & 45.0 \\
          & w/o GT & (2,12,13,15) &            &            & 41.0 \\
          \cline{2-6}
          & w/ GT & (2,11,12,15) & \checkmark & \checkmark & 53.4 \\
          & w/ GT & (2,11,12,15) & \checkmark &            & 46.7 \\
          & w/ GT & (2,11,12,15) &            &            & 41.3 \\
          \Xhline{1pt}
    \end{tabular}%
  }
  \label{tab:beamsearch-OT-RHM}%
\end{table}%

\paragraph{Beam search, OT and RHM.} To validate the effectiveness of adopting the pixel-level contrastive loss as the objective in beam search (Section~\ref{sec:beam-search}), we compare the results by performing beam search with and without using the ground truth annotations in the validation set. 
%
As shown in Table~\ref{tab:beamsearch-OT-RHM}, beam search using only the pixel-level contrastive loss performs effectively in our method, as it gives nearly the same hyperpixel selection as the standard beam search algorithm (51.0\% vs. 53.4\%). 
%
This also reveals that the proposed pixel-level contrastive loss, as an unsupervised surrogate loss, facilitates learning effective fine-grained feature representations for cross-instance matching.  
%
%
In addition, we also study the effectiveness of OT and RHM. 
The results in Table~\ref{tab:beamsearch-OT-RHM} show that both OT and RHM effectively facilitate the matching process. 
More interestingly, even without using these two post-processing steps, our method achieves significant performance gain compared to the baseline approach (41.0\% vs.\ 18.8\% w/o GT).
Fig.~\ref{fig:vis-comparison} shows some semantic correspondences by the evaluated methods.
We note that OT and RHM regularize the correspondence by avoiding many-to-one mapping and by making all keypoints' correspondences geometrically consistent.
Even without the post-processing steps (Fig.~\ref{fig:vis-comparison}, Raw), our method can determine more geometrically consistent alignments than those by SCOT.
These results demonstrate that the proposed pixel-level contrastive loss is effective in finding geometrically consistent matching by regularizing the learned representation.

\paragraph{Information entropy regularization.} 
%
We validate the effectiveness of the information entropy loss~\eqref{eq:information-entropy-loss}. 
As shown in Table~\ref{tab:comparison-sota}, the information entropy regularization effectively improves the performance on the PF-WILLOW dataset.
However, it does not have the same effect on the PF-PASCAL dataset.
We note that our method also significantly outperforms Weakalign~\citep{Rocco2018EndtoEndWS} (Table~\ref{tab:comparison-sota}), which solely utilizes the soft-inlier loss (a variant of entropy formulation).

\begin{figure*}[tp]
\centering
\begin{tabular}{cc}
   \begin{tabular}{*{2}{p{0.11\linewidth}<{\centering}}}
      $I'_0$ & $I_1$
   \end{tabular} &
   \begin{tabular}{*{5}{p{0.10\linewidth}<{\centering}}}
      $I'_0$ & $I_0$ (cycle) & $I_0$ (direct) & $I_0$ (GT) & Attention
   \end{tabular} \\
   \includegraphics[width=0.25\textwidth]{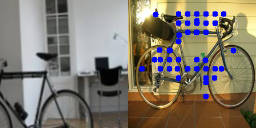}  & \includegraphics[width=0.625\textwidth]{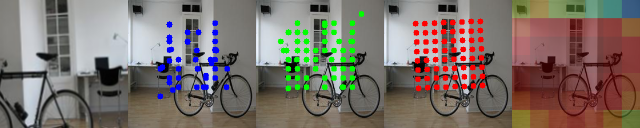}  \\
   \includegraphics[width=0.25\textwidth]{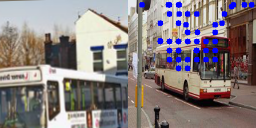}  & \includegraphics[width=0.625\textwidth]{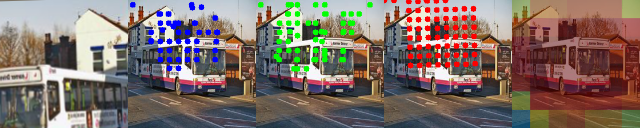}  \\
   \includegraphics[width=0.25\textwidth]{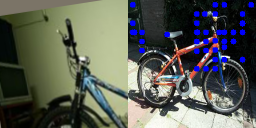}  & \includegraphics[width=0.625\textwidth]{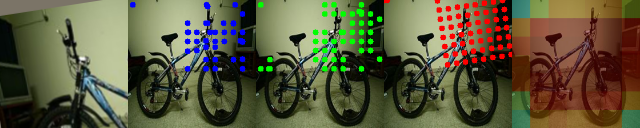}  \\
   \includegraphics[width=0.25\textwidth]{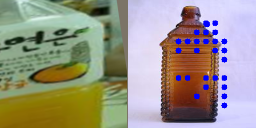}  & \includegraphics[width=0.625\textwidth]{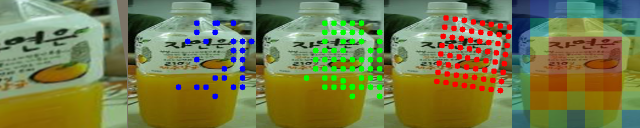}  \\
   \includegraphics[width=0.25\textwidth]{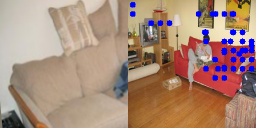}  & \includegraphics[width=0.625\textwidth]{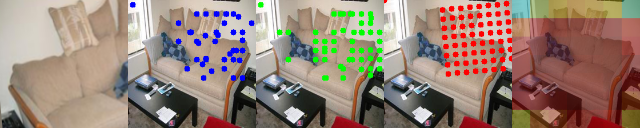}  \\
   \includegraphics[width=0.25\textwidth]{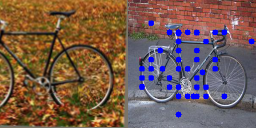}  & \includegraphics[width=0.625\textwidth]{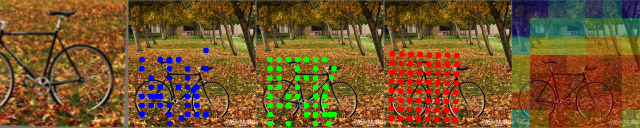}  \\
   \includegraphics[width=0.25\textwidth]{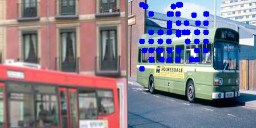}  & \includegraphics[width=0.625\textwidth]{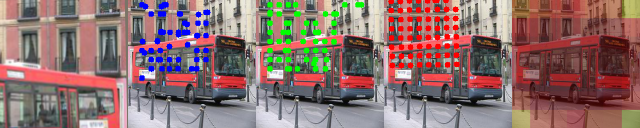}  \\
   \includegraphics[width=0.25\textwidth]{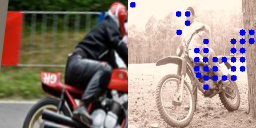}  & \includegraphics[width=0.625\textwidth]{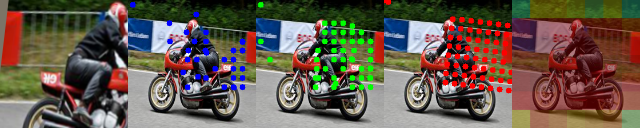}  \\
\end{tabular}
\caption{Visualization of pixel-level contrastive learning. The first two columns show correspondence predictions across different images (i.e., the blue dots in the second column denote the correspondence of each pixel in $I'_0$). The remaining five columns show correspondence predictions from the augmented patch images $I'_0$ to their source images $I_0$, where the red dots denote the correspondence ground truth (where $I'_0$ is cropped from $I_0$), blue or green dots indicate the correspondence prediction based on the cycle affinity matrix $\bar{A}_{0'0} = A_{0'1}A_{10}$ or affinity matrix $A_{0'0}$. The last column shows the self-attention map on which the randomly cropped augmented image $I'_0$ is based. Note that the colored dots are computed based on the feature map. Thus the number of colored dots is less than the number of pixels.}
\label{fig:vis-training}
\end{figure*}

\paragraph{Self-attention module.}
We evaluate the effectiveness of the self-attention module used for augmentation in the pixel-level contrastive learning method.  
As shown in Table~\ref{tab:self-attention}, the self-attention module improves the matching performance of our model. 
Without employing the self-attention module, background pixels may be included in the cropped patch image $I'_0$. 
However, there is no semantic correspondence for those background pixels in the target image $I_1$.
Taking the images in Fig.~\ref{fig:framework}(b) as an example, the pixels in the grass regions of $I_0$ should not be matched to those in the wall regions of $I_1$ since they do not belong to the same category. 
The self-attention module can help remove the false positive samples in the background to achieve better results. More visualization of the attention map can be found in Fig.~\ref{fig:vis-training}.

\begin{table}[tp]
    \centering
    \caption{Ablation study of the self-attention module on PF-PASCAL. We evaluate the performance at PCK@0.05.}
    \begin{tabular}{c|c|c}
        \Xhline{1pt}
         & Self-Attention & PCK@0.05 \\
        \Xhline{1pt}
        Ours & \checkmark & 51.0\\
        Ours (w/o attention) &  & 49.9 \\
        \Xhline{1pt}
    \end{tabular}
    \label{tab:self-attention}
\end{table}

\paragraph{Different layers of feature for training.}
We explore the effectiveness of features from different layers  for pixel-level contrastive learning. 
For example, the ResNet50 model contains 16 res-blocks. 
We classify features before the $7^\mathrm{th}$ block as low-level features, those after the $13^\mathrm{th}$ block as high-level features, and the rest as middle-level features. 
Since features from shallow layers usually entail low-level vision clues, e.g., color and texture, and high-level layer features are invariant to different local regions (i.e., via image-contrastive loss), we only evaluate the mid-level features. 
Table~\ref{tab:training-layer} shows that the features from layer 7 to 13 help achieve similar performance while the $10^\mathrm{th}$ layer yields the best features. 
%
Even so, we utilize the features following $13^\mathrm{th}$ layer for pixel-level contrastive learning in the other experiments to fairly compare with other state-of-the-art methods.
%

\begin{table}[tp]
  \centering
  \caption{Evaluation results of using features from different layers on PF-PASCAL. The first column denotes the residual block ID where the feature for pixel-level contrastive learning is extracted. The second column represents the hyperpixel used for testing our beam search algorithm. }
    \begin{tabular}{c|c|c}
        \Xhline{1pt}
        Block ID & Hyperpixel  & PCK@0.05 \\
        \Xhline{1pt}
        13    & (2,12,13,15) & 51.0 \\
        12    & (2,6,13,15)  & 49.1 \\
        10    & (2,6,13,15)  & 52.0 \\
        9     & (2,6,13,15)  & 51.3 \\
        7     & (2,6,13,15)  & 49.1 \\
        \Xhline{1pt}
    \end{tabular}%
  \label{tab:training-layer}%
\end{table}%

\paragraph{Differentiable OT, RHM, and concentration loss~\citep{Li2019JointtaskSL}.}
We observe that both OT and RHM effectively improve the model performance when used as post-processing steps in the testing pipeline. 
Motivated by this, we analyze whether they can improve the performance in our model when including them in the training phase. 
We implement both OT and RHM as differentiable layers without trainable parameters. 
The function of the differentiable OT layer is equivalent to the Hungarian algorithm~\citep{Munkres1957AlgorithmsFT}, classically used for bipartite matching. 
We apply one iteration of the Sinkhorn’s algorithm to the affinity matrix and obtain the transport matrix. 
Differentiable RHM aims to achieve geometrically consistent matching by re-weighting the matching scores and outputs the voting matrix. 
Here we insert the differentiable OT and RHM layers into the training pipeline and compute the pixel-level contrastive loss on either transport matrix (after OT) or voting matrix (after OT and RHM) instead of the affinity matrix. 
In addition, we adopt the concentration loss~\citep{Li2019JointtaskSL} to encourage more concentrated keypoints predictions for the patch image. 

By adopting differentiable OT and/or RHM layers, we observe an unstable training process. 
The results also reveal that the differentiable OT and RHM layers negatively affect the representation learning performance (Table~\ref{tab:Differentiable-OT-RHM-concentration}).
This is potentially caused by the distorted feature space.
%
%
The gradients become unstable because the differentiable OT is carried out by iterative matrix inversion.
%
%
We also observe that the matched pixels are already concentrated (see Fig.~\ref{fig:vis-comparison}), and thus adding the concentration regularization does not make significant differences. 

\begin{table}[tp]
  \centering
  \caption{Ablation study of differentiable OT, RHM, and concentration loss on PF-PASCAL. OT and RHM are used in training as differentiable layers. The concentration loss is employed as a regularization during training. The results below are evaluated at PCK\@0.05, and the bold number indicates the best performance.}
  \scalebox{1}{
    \begin{tabular}{c|c|c|c}
        \Xhline{1pt}
        OT &  RHM  &  Concentration  &  PCK@0.05 \\
        \Xhline{1pt}
            &       &       & {\bf 51.0} \\
        \checkmark & \checkmark &       & 32.4 \\
        \checkmark &       &       & 41.5 \\
            &       & \checkmark & 48.1 \\
        \Xhline{1pt}
    \end{tabular}%
   }
  \label{tab:Differentiable-OT-RHM-concentration}%
\end{table}%

\begin{figure*}[htp]
\centering
\def\picwidth{0.45\textwidth}
\begin{tabular}{cc}
 Prediction & Ground Truth  \\
\includegraphics[width=\picwidth]{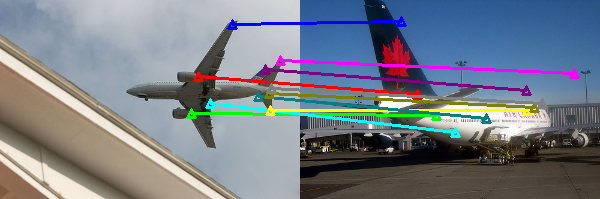} &
\includegraphics[width=\picwidth]{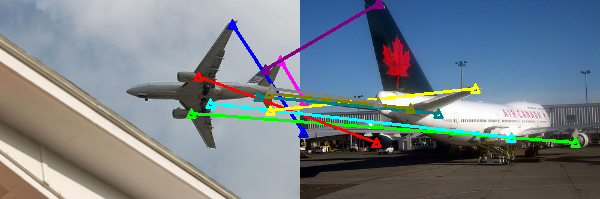} \\
 \includegraphics[width=\picwidth]{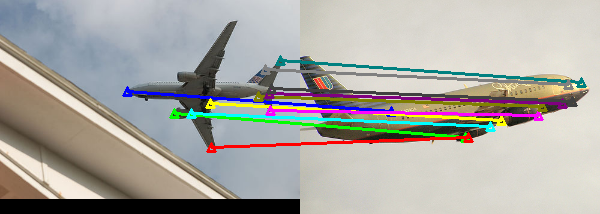} &
 \includegraphics[width=\picwidth]{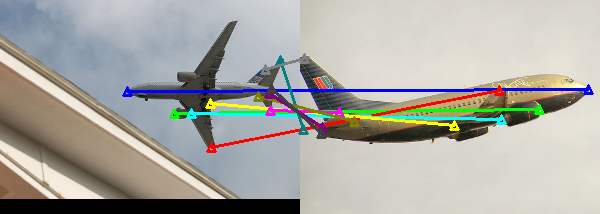} \\
\includegraphics[width=\picwidth]{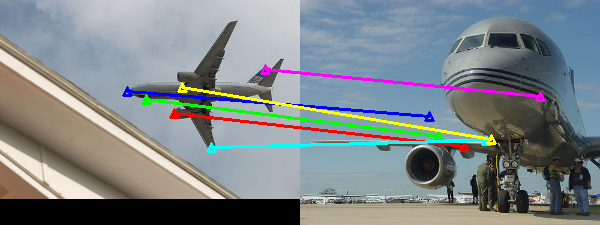} &
\includegraphics[width=\picwidth]{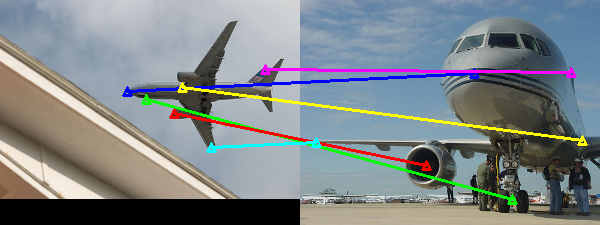} \\
\includegraphics[width=\picwidth]{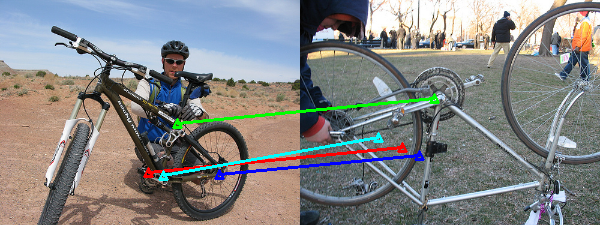} &
\includegraphics[width=\picwidth]{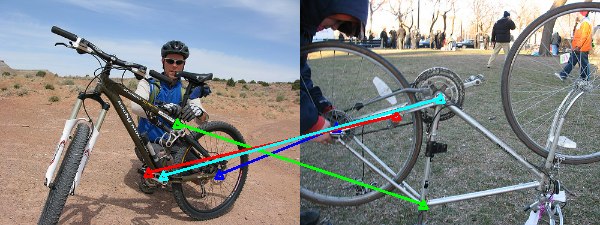} \\
\includegraphics[width=\picwidth]{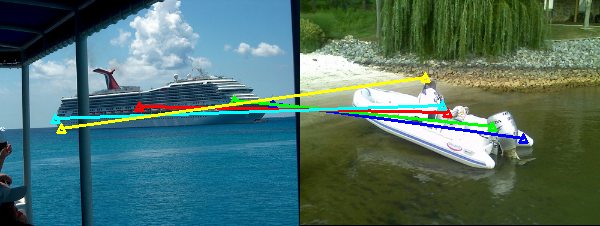} &
\includegraphics[width=\picwidth]{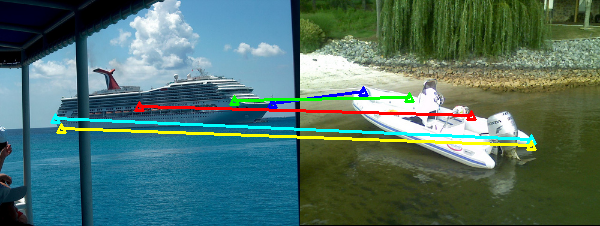} \\
\includegraphics[width=\picwidth]{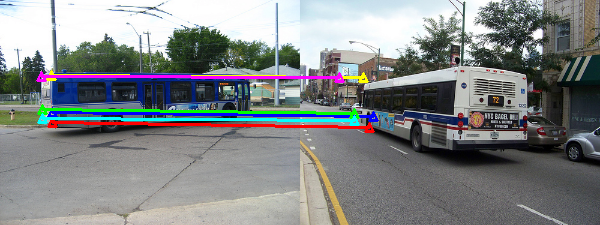} &
\includegraphics[width=\picwidth]{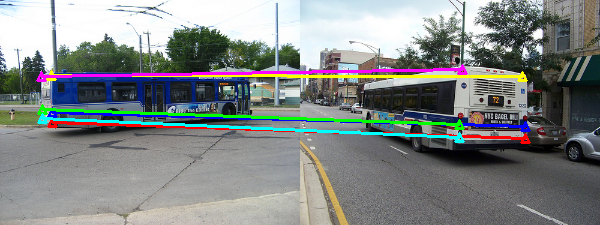} \\
\includegraphics[width=\picwidth]{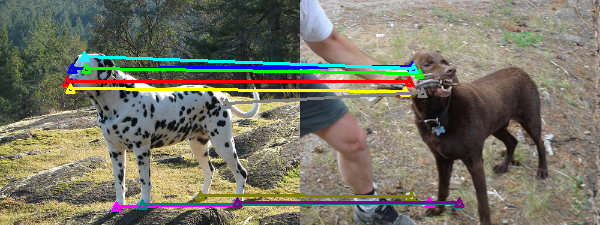} &
\includegraphics[width=\picwidth]{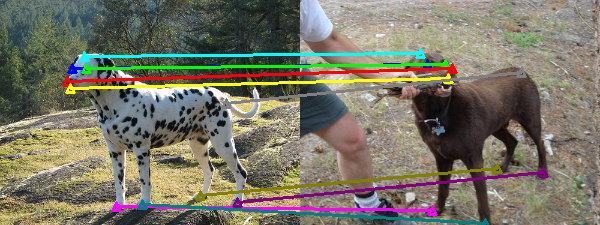} \\
\end{tabular}
\caption{Visualization of the correspondence predictions on SPair-71k. The left and right columns are the predictions of our method and the ground truth, respectively. Different keypoints matches are indicated by different colors. 
The predictions are based on the voting matrix by going through the process of OT and RHM. See Section~\ref{sec:discussion-limitation} for a detailed discussion.}
\label{fig:vis-spair}
\end{figure*}

\subsection{Discussion and Limitation}
\label{sec:discussion-limitation}

To better understand how the pixel-level contrastive loss function regularizes the feature representation, we visualize the correspondences given by the affinity matrices on the training dataset. 
For example, let $I_0$ and $I_1$ denote source and target images, and $I'_0$ the patch image randomly cropped from $I_0$.
Fig.~\ref{fig:vis-training} shows that the cycle constraint can help the cross-instance correspondence prediction of $I'_0$ onto $I_1$ by enforcing the correspondence predictions of $I'_0$ on $I_0$ to be the same as the ground truth. 
For example, the patch image $I'_0$ in the fourth row is cropped from the right part of a juice bottle. 
%
Interestingly, the correspondence prediction on $I_1$ (the second column) is also on the right side of a different bottle.
We observe a similar phenomenon in the other rows. 
These results demonstrate that multi-level contrastive learning helps focus on the contextual semantic information and learn spatially aware pixel-level feature representations. 

\revise{Numerous image-level contrastive learning methods have been proposed for object recognition and related tasks in recent years. 
%
However, it has not been exploited for learning correspondences especially on the semantic level. 
%
We show that image-level contrastive learning facilitates object-centric attention, which in turn helps finding semantic correspondence. 
%
Our approach is in direct contrast to existing semantic correspondence methods that mainly use ImageNet pretrained models for feature extraction.
%
%
}

\revise{Moreover, the proposed cross-instance pixel-level contrastive learning is specifically designed for semantic correspondence.
%
Existing pixel-level contrastive learning approaches~\citep{Wang2020DenseCL,Pinheiro2020UnsupervisedLO,Jabri2020SpaceTimeCA} are direct extensions of image-level instance contrastive learning (e.g., MoCo~\citep{He2020MomentumCF}). 
As such, they compute contrastive losses between different views of the same instance, e.g., either via self-augmented images~\citep{Wang2020DenseCL,Pinheiro2020UnsupervisedLO} or from sequences of the same instance~\citep{Jabri2020SpaceTimeCA,xie2021propagate}. 
Different from existing contrastive learning approaches, the proposed cross-instance pixel contrastive learning method leverages different images of the same category via cycle consistence.
This is not a straightforward extension of the aforementioned approaches, especially when we do not leverage any pixel-level ground truth for constructing positive/negative sample as opposed to the scheme in \citep{kang2020pixel}.
The proposed method aims to learn category-level mid-level representations, which is more challenging than the goals of existing approaches.
}

\begin{table}[tp]
  \centering
  \caption{Evaluation results on the SPair-71k dataset. Numbers in bold indicate the best performance.}
  \scalebox{0.9}{
    \begin{tabular}{l|c|c|c}
	\Xhline{1pt}
    \multicolumn{1}{c|}{\multirow{2}[0]{*}{Methods}} & \multicolumn{3}{c}{SPair-71k ($\alpha_{bbox}$)} \\
	\cline{2-4}
          & \multicolumn{1}{l|}{PCK@0.05} & \multicolumn{1}{l|}{PCK@0.10} & \multicolumn{1}{l}{PCK@0.15} \\
 	 \Xhline{1pt}
 	 \revise{HPF}~\citep{Min2019HyperpixelFS} & \revise{5.50} & \revise{14.20}  & \revise{22.70}  \\
     DHPF~\citep{Min2020LearningTC} & 5.03 & 13.71 & 22.58 \\
     SCOT~\citep{Liu2020SemanticCA} & 6.20 & 15.60 & 24.60 \\
     Ours & {\bf 6.30} & {\bf 15.80} & {\bf 25.20} \\
     \Xhline{1pt}
    \end{tabular}%
    }
  \label{tab:spair-comparison}%
\end{table}%

\begin{table}[tp]
  \centering
  \caption{\revise{Evaluation results on the SPair-71k dataset by using different percentages of pixel correspondence ground truth. For example, 5\% GT means that 5\% of training image pairs are provided with their pixel correspondence ground truth.}}
  \scalebox{1}{
    \begin{tabular}{l|c|c|c}
	\Xhline{1pt}
    \multicolumn{1}{c|}{\multirow{2}[0]{*}{Methods}} & \multicolumn{3}{c}{SPair-71k ($\alpha_{bbox}$)} \\
	\cline{2-4}
          & \multicolumn{1}{l|}{PCK@0.05} & \multicolumn{1}{l|}{PCK@0.10} & \multicolumn{1}{l}{PCK@0.15} \\
 	 \Xhline{1pt}
     Ours (0\% GT)   & 6.30  &  15.80 & 25.20 \\
     Ours (5\% GT)   & 6.58  &  16.03 & 25.41 \\
     Ours (10\% GT)  & 8.51  &  19.91 & 30.20 \\
     Ours (15\% GT)  & 9.75  &  22.49 & 33.56 \\
     Ours (100\% GT) & 11.57 &  25.83 & 37.60 \\
     \Xhline{1pt}
    \end{tabular}%
    }
  \label{tab:spair-comparison-semi}%
\end{table}%

We further compare our method with other baselines on the SPair-71k dataset~\citep{min2019spair}, which is a more challenging large-scale dataset. 
We use the self-supervised pretrained network from MoCo~\citep{He2020MomentumCF} as initialization for all methods. 
%
%
As shown in Table~\ref{tab:spair-comparison}, our method performs favorably against the HPF~\citep{Min2019HyperpixelFS}, DHPF~\citep{Min2020LearningTC} and SCOT~\citep{Liu2020SemanticCA} approaches.

\revise{
However, the results from the proposed self-supervised models 
are slightly worse than those by the supervised methods. 
To better understand the algorithmic performance, we use the self-supervised pretrained model as initialization and utilize some labeled image pairs with pixel correspondence ground truth for training. 
As shown in Table~\ref{tab:spair-comparison-semi}, leveraging pixel correspondence ground truth can improve model performance. }

We note that the SPair-71 dataset is significantly more challenging than the PF-PASCAL and PF-WILLOW databases, as there exist large pose variations, viewpoint changes, scale differences, and occlusions in it.
Fig.~\ref{fig:vis-spair} presents some of those challenging cases where our method fails to obtain a good result. 
There are three directions to improve our proposed method. 
First, self-supervised pretrained feature extractor needs to be improved to provide more robust feature representations by employing augmentations that can synthesize images with larger scale and viewpoint changes.
Second, we may need to combine the cross cycle consistency regularization with a keypoint detector to learn more fine-grained correspondences. 
\revise{Third, the image-level contrastive learning could be improved by including more similar instances from different images to construct positive samples.}
%
Our future work will focus on addressing these issues.  


\section{Conclusion}

%
We pose a new task to learn semantic correspondence without relying on supervised ImageNet pretrained models or ground truth annotations from the validation set. 
In the proposed method, we develop a multi-level contrastive learning framework where the image-level contrastive learning module generates object-level discriminative representations, and the pixel-level contrastive learning method constructs fine-grained representations to infer dense semantic correspondence. 
%
%
The pixel-level contrastive learning module is realized through the proposed cross-instance cycle consistency regularization, where we leverage different objects of the same category in different images without knowing their dense correspondence labels.
This is different from existing approaches where correspondences are constructed by augmenting the same object in images or video sequences. 
Experimental results on the PF-PASCAL, PF-WILLOW, and SPair-71k datasets demonstrate the effectiveness of the proposed method over the state-of-the-art schemes for semantic correspondence.

\begin{acknowledgements}
T. Xiao and M.-H. Yang are supported in part by NSF CAREER grant 1149783.
\end{acknowledgements}


%
%

\bibliographystyle{spbasic}      
\bibliography{egbib}   


\end{document}